\documentclass[conference]{IEEEtran}
\IEEEoverridecommandlockouts
\usepackage{cite}
\usepackage{amsmath,amssymb,amsfonts}
\usepackage{algorithmic}
\usepackage{graphicx}
\usepackage{textcomp}
\usepackage{multirow}
\usepackage{multicol}
\usepackage{booktabs}
\usepackage[table]{xcolor}
\usepackage{hyperref}
\newcommand{\tabincell}[2]{\begin{tabular}
{@{}#1@{}}#2\end{tabular}}
\usepackage[ruled,vlined]{algorithm2e}
\def\BibTeX{{\rm B\kern-.05em{\sc i\kern-.025em b}\kern-.08em
    T\kern-.1667em\lower.7ex\hbox{E}\kern-.125emX}}
\begin{document}

\title{Deep Reinforcement Learning for Imbalanced Classification
}

\author{\IEEEauthorblockN{ Enlu Lin\textsuperscript{1} $\qquad$    Qiong Chen\textsuperscript{2,*}  $\qquad$   Xiaoming Qi\textsuperscript{3}}
\IEEEauthorblockA{\textit{School of Computer Science and
Engineering} \\
\textit{South China University of Technology}\\
Guangzhou, China \\
linenus@outlook.com $\qquad$ csqchen@scut.edu.cn $\qquad$ qxmscut@126.com}
}

\maketitle

\begin{abstract}
Data in real-world application often exhibit skewed class distribution which poses an intense challenge for machine learning. Conventional classification algorithms are not effective in the case of imbalanced data distribution, and may fail when the data distribution is highly imbalanced. To address this issue, we propose a general imbalanced classification model based on deep reinforcement learning. We formulate the classification problem as a sequential decision-making process and solve it by deep Q-learning network. The agent performs a classification action on one sample at each time step, and the environment evaluates the classification action and returns a reward to the agent. The reward from minority class sample is larger so the agent is more sensitive to the minority class. The agent finally finds an optimal classification policy in imbalanced data under the guidance of specific reward function and beneficial learning environment. Experiments show that our proposed model outperforms the other imbalanced classification algorithms, and it can identify more minority samples and has great classification performance.
\end{abstract}

\begin{IEEEkeywords}
imbalanced classification, deep reinforcement learning, reward function, classification policy
\end{IEEEkeywords}

\section{Introduction}
Imbalanced data classification has been widely researched in the field of machine learning\cite{japkowicz2002class,weiss2004mining,he2008learning}. In some real-world classification problems, such as abnormal detection, disease diagnosis, risk behavior recognition, etc., the distribution of data across different classes is highly skewed. The instances in one class (e.g., cancer) can be 1000 times less than that in another class (e.g., healthy patient). Most machine learning algorithms are suitable for balanced training data set. When facing imbalanced scenarios, these models often provide a good recognition rates to the majority instances, whereas the minority instances are distorted. The instances in minority class are difficult to detect because of their infrequency and casualness; however, misclassifying minority class instances can result in heavy costs.  

A range of imbalanced data classification algorithms have been developed during the past two decades. The methods to tackle these issues are mainly divided into two groups\cite{haixiang2017learning}: the data level and the algorithmic level. The former group modifies the collection of instances to balance the class distribution by re-sampling the training data, which often represents as different types of data manipulation techniques. The latter group modifies the existing learners to alleviate their bias towards majority class, which often assigns higher misclassification cost to the minority class. However, with the rapid developments of big data, a large amount of complex data with high imbalanced ratio is generated which brings an enormous challenge in imbalanced data classification. Conventional methods are inadequate to cope with more and more complex data so that novel deep learning approaches are increasingly popular.  

In recent years, deep reinforcement learning has been successfully applied to computer games, robots controlling, recommendation systems\cite{mnih2013playing,gu2017deep,zhao2017deep} and so on. For classification problems, deep reinforcement learning has served in eliminating noisy data and learning better features, which made a great improvement in classification performance. However, there has been little research work on applying deep reinforcement learning to imbalanced data learning. In fact, deep reinforcement learning is ideally suitable for imbalanced data learning as its learning mechanism and specific reward function are easy to pay more attention to minority class by giving higher reward or penalty.  

A deep $Q$-learning network (DQN) based model for imbalanced data classification is proposed in this paper. In our model, the imbalanced classification problem is regarded as a guessing game which can be decomposed into a sequential decision-making process. At each time step, the agent receives an environment state which is represented by a training sample and then performs a classification action under the guidance of a policy. If the agent performs a correct classification action it will be given a positive reward, otherwise, it will be given a negative reward. The reward from minority class is higher than that of majority class. The goal of the agent is to obtain as more cumulative rewards as possible during the process of sequential decision-making, that is, to correctly recognize the samples as much as possible.   

The contributions of this paper are summarized as follows: 1) Formulate the classification problem as a sequential decision-making process and propose a deep reinforcement learning framework for imbalanced data classification. 2) Design and implement the DQN based imbalanced classification model DQNimb, which mainly includes building the simulation environment, defining the interaction rules between agent and environment, and designing the specific reward function. 3) Study the performance of our model through experiments and compare with the other methods of imbalanced data learning.   

The rest of this paper is organized as follows: The second section introduces the research methodology of imbalanced data classification and the  applications of deep reinforcement learning for classification problems. The third section elaborates the proposed model and analyzes it theoretically. The fourth section shows the experimental results and evaluates the performance of our method compared with the other methods. The last section summarizes the work of this paper and looks forward to the future work.

\section{Related Work}

\subsection{Imbalanced data classification}
The previous research work in imbalanced data classification concentrate mainly on two levels: the data level\cite{drummond2003c4,han2005borderline,mani2003knn,batista2004study} and the algorithmic level\cite{veropoulos1999controlling,wu2005kba,tang2009svms,zadrozny2001learning,zadrozny2003cost,zhou2006training,krawczyk2015cost,chen2006decision,yu2016odoc,ting2000comparative}. Data level methods aim to balance the class distribution by manipulating the training samples, including over-sampling minority class, under-sampling majority class and the combinations of the two above methods\cite{batista2004study}. SMOTE is a well-known over-sampling method, which generates new samples by linear interpolation between adjacent minority samples\cite{han2005borderline}. NearMiss is a typical under-sample method based on the nearest neighbor algorithm\cite{mani2003knn}. However, over-sampling can potentially lead to overfitting while under-sampling may lose valuable information on the majority class. The algorithmic level methods aim to lift the importance of minority class by improving the existing algorithms, including cost-sensitive learning, ensemble learning, and decision threshold adjustment. The cost-sensitive learning methods assign various misclassification costs to different classes by modifying the loss function, in which the misclassification cost of minority class is higher than that of majority class. The ensemble learning based methods train multiple individual sub-classifiers, and then use voting or combining to get better results. The threshold-adjustment methods train the classifier in original imbalanced data and change the decision threshold in test time. A number of deep learning based methods have recently been proposed for imbalanced data classification\cite{wang2016training,huang2016learning,yan2015deep,khan2018cost,dong2018imbalanced}. Wang \emph{et al}.\cite{wang2016training} proposed a new loss function in deep neural network which can capture classification errors from both majority class and minority class equally. Huang \emph{et al}.\cite{huang2016learning} studied a method that learns more discriminative feature of imbalanced data by maintaining both inter-cluster and inter-class margins. Yan \emph{et al}.\cite{yan2015deep} used a bootstrapping sampling algorithm which ensures the training data in each mini-batch for convolutional neural network is balanced. A method to  optimize the network parameters and the class-sensitive costs jointly was presented in \cite{khan2018cost}. In \cite{dong2018imbalanced} Dong \emph{et al}. mined hard samples in minority classes and improved the algorithm by batch-wise optimization with Class Rectification Loss function.  

\subsection{Reinforcement learning for classification problem}
Deep reinforcement learning has recently achieved excellent results in classification tasks as it can assist classifiers to learn advantageous features or select high-quality instances from noisy data. In \cite{wiering2011reinforcement}, the classification task was constructed into a sequential decision-making process, which uses multiple agents to interact with the environment to learn the optimal classification policy. However, the intricate simulation between agents and environment caused extremely high time complexity.  Feng \emph{et al}.\cite{feng2018reinforcement} proposed a deep reinforcement learning based model to learn the relationship classification in noisy text data. The model is divided into instance selector and relational classifier. The instance selector selects high-quality sentence from noisy data under the guidance of agent while the relational classifier learns better performance from selected clean data and feeds back a delayed reward to the instance selector. The model finally obtains a better classifier and high-quality data set. The work in \cite{zhang2018learning,liu2018deep,zhao2017deeprl,janisch2017classification} utilized deep reinforcement learning to learn advantageous features of training data in their respective applications. In general, the advantageous features improve the classifier while the better classifier feeds back a higher reward which encourages the agent to select more advantageous features. Martinez \emph{et al}.\cite{martinez2018deep} proposed a deep reinforcement learning framework for time series data classification in which the definition of specific reward function and the Markov process are clearly formulated. Research in imbalanced data classification with reinforcement learning was quite limited. In \cite{abdi2014ensemble} an ensemble pruning method was presented that selected the best sub-classifiers by using reinforcement learning. However, this method was merely suitable for traditional small dataset because it was inefficient to select classifiers when there were plenty of sub-classifiers. In this paper, we propose deep $Q$-network based model for imbalanced classification which is efficient in complex high-dimensional data such as image or text and has a good performance compared to the other imbalanced classification methods.  

\section{Methodology}
\subsection{Imbalanced Classification Markov Decision Process}
\begin{figure}
    \centering
    \includegraphics[width=8cm]{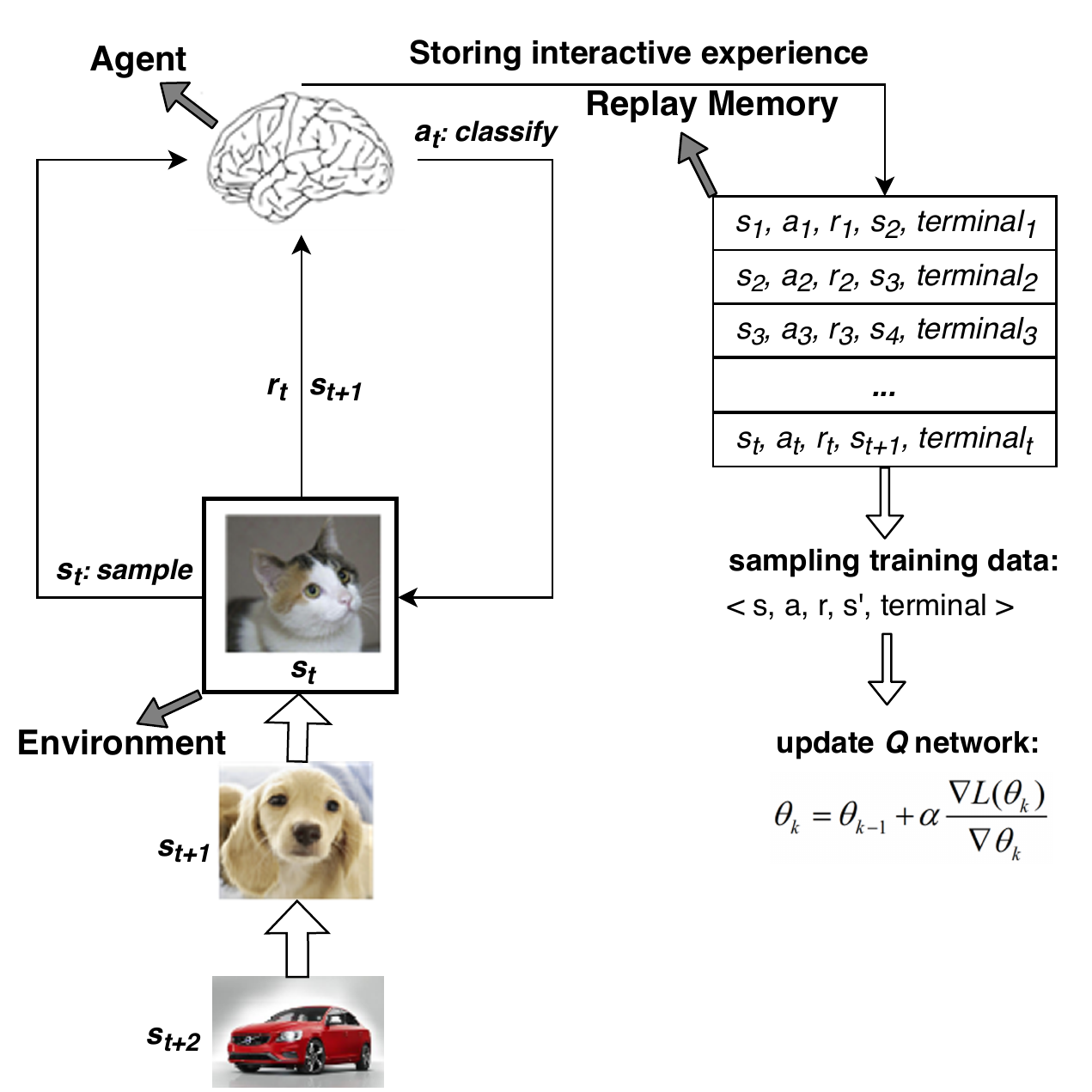}
    \caption{Overall process of ICMDP.}
    \label{fig1}
\end{figure}

Reinforcement learning algorithms that incorporate deep learning have defeated world champions at the game of Go as well as human experts playing numerous Atari video games. Now we regard classification problem as a guessing game, the agent receives a sample at each time step and guesses (classifies) which category the sample belongs to, and then the environment returns it an immediate reward and the next sample, as shown in Fig.\ref{fig1}. A positive reward is given to the agent by the environment when the agent correctly guesses the category of sample, otherwise a negative reward is given to the agent. When the agent learns an optimal behavior from its interaction with environment to get the maximum accumulative rewards, it can correctly classify samples as much as possible.  

Now we formalize the Imbalanced Classification Markov Decision Process (ICMDP) framework which decomposes imbalanced data classification task into a sequential decision-making problem. Assume that the imbalanced training data set is $D=\lbrace(x_1,l_1),(x_2,l_2),...,(x_n,l_n)\rbrace$ where $x_i$ is the ith sample and $l_i$ is the label of the ith sample. We propose to train a classifier as an agent evolving in ICMDP where:  
\begin{itemize}
\item \textbf{State} $\mathcal{S}$: The state of environment is determined by the training sample. At the beginning of training, the agent receives the first sample $x_1$ as its initial state $s_1$. The state $s_t$ of environment at each time step corresponds to the sample $x_t$. When the new episode begins, environment shuffles the order of samples in training data set. 
\item \textbf{Action} $\mathcal{A}$: The action of agent is associated with the label of training data set. The action $a_t$ taken by agent is to predict a class label. For binary classification problem, $\mathcal{A}=\lbrace 0,1\rbrace$ where 0 represents the minority class and 1 represents the majority class.
\item \textbf{Reward} $\mathcal{R}$: A reward $r_t$ is the feedback from environment by which we measure the success or failure of an agent's actions. In order to guide the agent to learn the optimal classification policy in imbalanced data,  the absolute reward value of sample in minority class is higher than that in majority class. That is, when the agent correctly or incorrectly recognizes minority class sample, the environment feedback agent a larger reward or punishment. 
\item \textbf{Transition probability} $\mathcal{P}$: Transition probability $p(s_{t+1}|s_t,a_t)$ in ICMDP is deterministic. The agent moves from the current state $s_t$ to the next state $s_{t+1}$ according to the order of samples in the training data set.  
\item \textbf{Discount factor} $\gamma$: $\gamma\in[0,1]$ is to balance the immediate and future reward.  
\item \textbf{Episode}: Episode in reinforcement learning is a transition trajectory from the initial state to the terminal state $\lbrace s_1,a_1,r_1,s_2,a_2,r_2,...,s_t,a_t,r_t\rbrace$. An episode ends when all samples in training data set are classified or when the agent misclassifies the sample from minority class.  
\item \textbf{Policy} $\pi_\theta$: The policy $\pi_\theta$ is a mapping function $\pi:\mathcal{S}\rightarrow \mathcal{A}$ where $\pi_\theta(s_t)$ denotes the action $a_t$ performed by agent in state $s_t$.  The policy $\pi_\theta$ in ICMDP can be considered as a classifier with the parameter $\theta$.   
\end{itemize}  

With the definitions and notations above, the imbalanced classification problem is formally defined as to find an optimal classification policy $\pi^\ast:\mathcal{S}\rightarrow \mathcal{A}$, which maximized the cumulative rewards in ICMDP.  

\subsection{Reward function for imbalanced data classification}
The minority class samples are difficult to be identified correctly in imbalance data set. In order to better recognize the minority class samples, the algorithm should be more sensitive to the minority class. A large reward or punishment is returned to agent when it meets a minority sample. The reward function is defined as follows:
\begin{equation}\label{eq.1}
R(s_t,a_t,l_t)=
\begin{cases}
+1,  & \text{$a_t=l_t$ and $s_t\in D_P$} \\
-1,  & \text{$a_t\neq l_t$ and $s_t\in D_P$} \\
\lambda,  & \text{$a_t=l_t$ and $s_t\in D_N$} \\
-\lambda, & \text{$a_t\neq l_t$ and $s_t\in D_N$}
\end{cases}    
\end{equation}
where $\lambda \in [0,1]$, $D_P$ is minority class sample set, $D_N$ is majority class sample set, $l_t$ is the class label of the sample in state $s_t$. Let the reward value be $1/-1$ when the agent correctly/incorrectly classifies a minority class sample, be $\lambda/-\lambda$ when the agent correctly/incorrectly classifies a majority class sample.

The value of reward function is the prediction cost of agent. For imbalanced data set ($\lambda<1$), the prediction cost values of minority class are higher than that of majority class. If the class distribution of training data set is balanced, then $\lambda=1$, the prediction cost values are the same for all classes. In fact, $\lambda$ is a trade-off parameter to adjust the importance of majority class. Our model achieves the best performance in experiment when $\lambda$ is equal to the imbalanced ratio $\rho=\frac{|D_P|}{|D_N|}$. We will discuss it in Section~\ref{sec:IRF}.

\subsection{DQN based imbalanced classification algorithm}  
  
\subsubsection{Deep $Q$-learning for ICMDP}
In ICMDP, the classification policy $\pi$ is a function which receives a sample and return the probabilities of all labels.  
\begin{equation}\label{eq.2}
\pi(a|s)=P(a_t=a|s_t=s)   
\end{equation}

The classifier agent's goal is to correctly recognize the sample of training data as much as possible. As the classifier agent can get a positive reward when it correctly recognizes a sample, thus it can achieve its goal by maximizing the cumulative rewards $g_t$:
\begin{equation}\label{eq.3}
g_t=\sum_{k=0}^{\infty}\gamma^k r_{t+k}    
\end{equation}

In reinforcement learning, there is a function that calculates the quality of a state-action combination, called the $Q$ function:
\begin{equation}\label{eq.4}
Q^\pi(s,a)=E_\pi[g_t|s_t=s,a_t=a]    
\end{equation}

According to the Bellman equation\cite{dixit1990optimization}, the $Q$  function can be expressed as:
\begin{equation}\label{eq.5}
Q^\pi(s,a)=E_\pi[r_t+\gamma Q^\pi(s_{t+1},a_{t+1})|s_t=s,a_t=a]    
\end{equation}

The classifier agent can maximize the cumulative rewards by solving the optimal $Q^\ast$ function, and the greedy policy under the optimal $Q^\ast$ function is the optimal classification policy $\pi^\ast$ for ICMDP.
\begin{equation}\label{eq.6}
\pi^\ast(a|s)=
\begin{cases}
1,  & \text{if $a=\arg\max_a Q^\ast(s,a)$} \\
0,  & \text{else} \\
\end{cases}    
\end{equation}

Substituting \eqref{eq.6} into \eqref{eq.5}, the optimal $Q^\ast$ function can be shown as:  
\begin{equation}\label{eq.7}
Q^\ast(s,a)=E_\pi[r_t+\gamma\max_{a}Q^\ast(s_{t+1},a_{t+1})|s_t=s,a_t=a]
\end{equation}

In the low-dimensional finite state space, $Q$ functions are recorded by a table. However, in the high-dimensional continuous state space, $Q$ functions cannot be resolved until deep $Q$-learning algorithm was proposed, which fits the $Q$ function with a deep neural network. In deep $Q$-learning algorithm, the interaction data $(s,a,r,s')$ obtained from \eqref{eq.7} are stored in the experience replay memory $M$. The agent randomly samples a mini-batch of transitions $B$ from $M$ and performs a gradient descent step on the Deep $Q$ network according to the loss function as follow:
\begin{equation}\label{eq.8}
L(\theta_k)=\sum_{(s,a,r,s')\in B}(y-Q(s,a;\theta_k))^2    
\end{equation}
where $y$ is the target estimate of the $Q$ function, the expression of $y$ is:
\begin{equation}\label{eq.9}
y=
\begin{cases}
r,  & \text{$terminal$=True} \\
r+\gamma\max_{a'}Q(s',a';\theta_{k-1}),  & \text{$terminal$=False} \\
\end{cases}    
\end{equation}
where $s'$ is the next state of $s$, $a'$ is the action performed by agent in state $s'$. 

The derivative of loss function \eqref{eq.8} with respect to $\theta$ is:
\begin{equation}\label{eq.10}
\frac{\nabla L(\theta_k)}{\nabla\theta_k}=-2\sum_{(s,a,r,s')\in B}(y-Q(s,a;\theta_k))\frac{\nabla Q(s,a;\theta_k)}{\nabla\theta_k}    
\end{equation}
 
Now we can obtain the optimal $Q^\ast$ function by minimizing the loss function \eqref{eq.8}, the greedy policy \eqref{eq.6} under the optimal $Q^\ast$ function will get the maximum cumulative rewards. So the optimal classification policy $\pi^\ast:\mathcal{S}\rightarrow \mathcal{A}$ for ICMDP is achieved.

\subsubsection{Influence of reward function}
In imbalanced data, the trained $Q$ network will be biased toward the majority class. However, due to the aforementioned reward function \eqref{eq.1}, it assigns different rewards for different classes and ultimately makes the samples from different classes have the same impact on $Q$ network.  

Suppose the positive and negative samples are denoted as $s^+$ and $s^-$, their target $Q$ values are represented as $y^+$ and $y^-$. According to \eqref{eq.1} and \eqref{eq.9}, the target $Q$ value of positive and negative samples is expressed as:
\begin{equation}\label{eq.12}
y^+=
\begin{cases}
(-1)^{1-I(a=l)}, & \text{$terminal$=True} \\
(-1)^{1-I(a=l)}+\gamma\max_{a'}Q(s',a'), & \text{$terminal$=False} \\
\end{cases}    
\end{equation}
\begin{equation}\label{eq.13}
y^-=
\begin{cases}
(-1)^{1-I(a=l)}\lambda, & \text{$terminal$=True} \\
(-1)^{1-I(a=l)}\lambda+\gamma\max_{a'}Q(s',a'), & \text{$terminal$=False} \\
\end{cases}    
\end{equation}
where $I(x)$ is an indicator function. 

Rewrite the loss function $L(\theta_k)$ of $Q$ network to the form of the sum of positive class loss function $L_+(\theta_k)$ and negative class loss function $L_-(\theta_k)$.  The derivative of $L_+(\theta_k)$ and $L_-(\theta_k)$ is shown as follows:
\begin{equation}\label{eq.14}
\frac{\nabla L_+(\theta_k)}{\nabla\theta_k}=-2\sum\nolimits_{i=1}^P\left(y_i^+-Q(s_i^+,a_i;\theta_k)\right)\frac{\nabla Q(s_i^+,a_i;\theta_k)}{\nabla\theta_k}    
\end{equation}

\begin{equation}\label{eq.15}
\frac{\nabla L_-(\theta_k)}{\nabla\theta_k}=-2\sum\nolimits_{j=1}^N\left(y_j^--Q(s_j^-,a_j;\theta_k)\right)\frac{\nabla Q(s_j^-,a_j;\theta_k)}{\nabla\theta_k}    
\end{equation}
where $P$ is the total number of the positive samples set, $N$ is the total number of the negative samples set.  

Substituting \eqref{eq.12} into \eqref{eq.14}, \eqref{eq.13} into  \eqref{eq.15} and adding the derivative of $L_+(\theta_k)$ and $L_-(\theta_k)$, then we get the following:
\begin{equation}\label{eq.16}
 \begin{split}
\frac{\nabla L(\theta_k)}{\nabla\theta_k}=&-2\sum\nolimits_{m=1}^{P+N}((1-t_m)\gamma\max_{a_m'}Q(s_m',a_m';\theta_{k-1})  \\
 &-Q(s_m,a_m;\theta_{k}))\frac{\nabla Q(s_m,a_m;\theta_k)}{\nabla\theta_k}\\
 &-2\sum\nolimits_{i=1}^P(-1)^{1-I(a_i=l_i)}\frac{\nabla Q(s_i,a_i;\theta_k)}{\nabla\theta_k}\\
 &-2\lambda\sum\nolimits_{j=1}^N(-1)^{1-I(a_j=l_j)}\frac{\nabla Q(s_j,a_j;\theta_k)}{\nabla\theta_k}
\end{split}   
\end{equation}  
where $t_m$=1 if $terminal$=True, otherwise $t_m$=0.

In \eqref{eq.16}, the second item relates to the minority class and the third item relates to the majority class. For imbalanced data set ($N>P$), if $\lambda=1$, the immediate rewards of the two classes are identical, the value of the third item is larger than that of the second item because the number of samples in majority class are much more than that in minority class. So the model is biased to the majority class. If $\lambda<1$, $\lambda$ can reduce the immediate rewards of negative samples and weakens their impact on the loss function of $Q$ network. What's more, the second item has the same value as the third item when $\lambda$ is equal to the imbalanced ratio $\rho$.

\begin{algorithm}\label{alg1}
    \caption{Training}
    \KwIn{Training data $D=\lbrace(x_1,l_1),(x_2,l_2),...,(x_T,l_T)\rbrace$. Episode number K.}
    Initialize experience replay memory $M$\\
    Randomly initialize parameters $\theta$\\
    Initialize simulation environments $\varepsilon$\\
    \For{$episode$ $k=1$ to K}
    {
        Shuffle the training data $D$\\
        Initialize state $s_1=x_1$\\
        \For{ $t=1$ to $T$ }
        {
            Choose an action based $\epsilon$-greedy policy:\\
            $a_t=\pi_\theta(s_t)$\\
            $r_t,terminal_t=STEP(a_t,l_t)$\\
            Set $s_{t+1}=x_{t+1}$\\
            Store $(s_t,a_t,r_t,s_{t+1},terminal_t)$ to $M$\\
            Randomly sample $(s_j,a_j,r_j,s_{j+1},terminal_j)$ from $M$\\
            Set $y_j=
                \begin{cases}
                r_j,  & \text{$terminal_j$=True} \\
                r_j+\gamma\max_{a'}Q(s_{j+1},a';\theta),  & \text{$terminal_j$=False} \\
                \end{cases}
            $\\
            Perform a gradient descent step on $L(\theta)$ w.r.t. $\theta$:
                $L(\theta)=(y_j-Q(s_j,a_j;\theta))^2$\\
            \If {$terminal_t$=True}
                {break}
        }
    }
\end{algorithm}

\SetKwProg{Fn}{Function}{}{}
\begin{algorithm}\label{alg2}
    \caption{Environment simulation}
    $D_P$ represents the minority class sample set.\\
    \Fn{STEP($a_t\in \mathcal{A}$, $l_t\in L$)}
    {
        Initialize $terminal_t$=False\\
        \eIf {$s_t\in D_P$}
        {
            \eIf{$a_t=l_t$}
            {
                Set $r_t$=1\\
            }
            {
                Set $r_t$=-1\\
                $terminal_t$=True\\
            }
        }
        {
            \eIf{$a_t=l_t$}
            {
                Set $r_t=\lambda$\\
            }
            {
                Set $r_t=-\lambda$\\
            }
        }
        return $r_t,terminal_t$\\
    }
    
\end{algorithm}

\subsubsection{Training details}
 We construct the simulation environment according to the definition of ICMDP. The architecture of the $Q$ network depends on the complexity and amount of training data set. The input of the $Q$ network is consistent with the structure of training sample, and the number of outputs is equal to the number of sample categories. In fact, the $Q$ network is a neural network classifier without the final softmax layer. The training process of $Q$ network is described in Algorithm \ref{alg1}. In an episode, the agent uses the $\epsilon$-greedy policy to pick the action, and then obtains the reward from the environment through the $STEP$ function in Algorithm \ref{alg2}. The deep $Q$-learning algorithm will be running about 120000 iterations (updates of network parameters $\theta$). We save the parameters of the converged $Q$ network which plus a softmax layer can be regarded as a neural network classifier trained by imbalanced data.

\section{Experiment}

\subsection{Comparison Methods and Evaluation Metrics}
We compare our method DQNimb with five imbalanced data learning methods from the data level and the algorithmic level, including sampling techniques, and cost-sensitive learning methods and decision threshold adjustment method. A deep neural network trained with cross entropy loss function will be used as baseline in our experiments. The comparison methods are shown as follows:
\begin{itemize}
\item \textbf{DNN}: A method which trains the deep neural network using cross entropy loss function without any improvement strategy in imbalanced data set.
\item \textbf{ROS}: A re-sampling method to build a more balanced data set through over-sampling minority classes by random replication \cite{drummond2003c4}.
\item \textbf{RUS}: A re-sampling method to build a more balanced data set through under-sampling majority classes by random sample removal \cite{drummond2003c4}.
\item \textbf{MFE}: A method to improve the classification performance of deep neural network in imbalanced data sets by using mean false error loss function \cite{wang2016training}
\item \textbf{CSM}: A cost sensitive method which assigns greater misclassification cost to minority class and smaller cost to majority class in loss function\cite{zhou2006training}  
\item \textbf{DTA}: A method to train the deep neural network in imbalanced data and to adjust the model decision threshold in test time by incorporating the class prior probability\cite{chen2006decision}  
\end{itemize}  

In our experiment, to evaluate the classification performance in imbalanced data sets more reasonably, G-mean and F-measure metrics\cite{gu2009evaluation} which are popularly used in imbalanced data sets are adopted. G-mean is the geometric mean of sensitivity and precision: G-mean=$\sqrt{\frac{TP}{TP+FN}\times\frac{TN}{TN+FP}}$. F-measure represents a harmonic mean between recall and precision: F-measure=$\sqrt{\frac{TP}{TP+FN}\times\frac{TP}{TP+FP}}$. The higher the G-mean score and F-measure score are , the better the algorithm performs.

\begin{table}
\caption{Dataset of Experiments}
\label{tbl1}
\begin{tabular}{p{0.75cm}|p{1.1cm}|p{1.1cm}|p{0.8cm}|p{0.85cm}|p{0.7cm}|p{0.7cm}}
\hline
\multirow{2}{*}{Dataset}       & \multirow{2}{*}{\tabincell{c}{Dimension\\of sample}} & \multirow{2}{*}{\tabincell{c}{Imbalance\\ratio $\rho$}} & \multicolumn{2}{c|}{Training data}        & \multicolumn{2}{c}{Test data}                  \\ \cline{4-7} 
                               &                                      &                                     & Pt.nmb$^{\mathrm{a}}$ & Ng.nmb$^{\mathrm{b}}$       & Pt.nmb       & Ng.nmb       \\ \hline
\multirow{3}{*}{IMDB}          & \multirow{3}{*}{1*500}               & 10\%                                & 1250             & \multirow{3}{*}{12000} & \multirow{3}{*}{12500} & \multirow{3}{*}{12500} \\ \cline{3-4}
                               &                                      & 5\%                                 & 625              &                        &                        &                        \\ \cline{3-4}
                               &                                      & 2\%                                 & 250              &                        &                        &                        \\ \hline
\multirow{4}{*}{\tabincell{c}{Cifar\\-10(1)}}   & \multirow{8}{*}{32*32*3}             & 4\%                                 & 400              & \multirow{4}{*}{10000} & \multirow{4}{*}{1000}  & \multirow{4}{*}{2000}  \\ \cline{3-4}
                               &                                      & 2\%                                 & 200              &                        &                        &                        \\ \cline{3-4}
                               &                                      & 1\%                                 & 100              &                        &                        &                        \\ \cline{3-4}
                               &                                      & 0.5\%                               & 50               &                        &                        &                        \\ \cline{1-1} \cline{3-7} 
\multirow{4}{*}{\tabincell{c}{Cifar\\-10(2)}}   &                                      & 4\%                                 & 800              & \multirow{4}{*}{20000} & \multirow{4}{*}{1000}  & \multirow{4}{*}{4000}  \\ \cline{3-4}
                               &                                      & 2\%                                 & 400              &                        &                        &                        \\ \cline{3-4}
                               &                                      & 1\%                                 & 200              &                        &                        &                        \\ \cline{3-4}
                               &                                      & 0.5\%                               & 100              &                        &                        &                        \\ \hline
\multirow{4}{*}{\tabincell{c}{Fashion-\\Mnist(1)}} & \multirow{8}{*}{28*28*1}             & 4\%                                 & 480              & \multirow{4}{*}{12000} & \multirow{4}{*}{2000}  & \multirow{4}{*}{2000}  \\ \cline{3-4}
                               &                                      & 2\%                                 & 240              &                        &                        &                        \\ \cline{3-4}
                               &                                      & 1\%                                 & 120              &                        &                        &                        \\ \cline{3-4}
                               &                                      & 0.5\%                               & 60               &                        &                        &                        \\ \cline{1-1} \cline{3-7}
\multirow{4}{*}{\tabincell{c}{Fashion-\\Mnist(2)}} &              & 4\%                                 & 720              & \multirow{4}{*}{18000} & \multirow{4}{*}{3000}  & \multirow{4}{*}{3000}  \\ \cline{3-4}
                               &                                      & 2\%                                 & 360              &                        &                        &                        \\ \cline{3-4}
                               &                                      & 1\%                                 & 180              &                        &                        &                        \\ \cline{3-4}
                               &                                      & 0.5\%                               & 90               &                        &                        &                        \\ \hline
\multirow{4}{*}{Mnist}         & \multirow{4}{*}{28*28*1}             & 1\%                                 & 540              & \multirow{4}{*}{54042} & \multirow{4}{*}{1032}  & \multirow{4}{*}{8968}  \\ \cline{3-4}
                               &                                      & 0.2\%                               & 108              &                        &                        &                        \\ \cline{3-4}
                               &                                      & 0.1\%                               & 54               &                        &                        &                        \\ \cline{3-4}
                               &                                      & 0.05\%                              & 27               &                        &                        &                        \\ \hline
\multicolumn{7}{l}{$^{\mathrm{a}}$Number of Positive class samples. $^{\mathrm{b}}$Number of Negative class samples.}
\end{tabular}
\end{table}

\subsection{Dataset}
In this paper, we mainly study the binary imbalanced classification with deep reinforcement learning. We perform experiments on IMDB, Cifar-10, Mnist and Fashion-Minist. Our approach is evaluated on the deliberately imbalanced splits. The simulated datasets used for the experiments are shown in Table \ref{tbl1}.

\textbf{IMDB} is a text dataset, which contains 50000 movies reviews labeled by sentiment (positive/negative). Reviews have been preprocessed, and each review is encoded as a sequence of word indexes. The standard train/test split for each class is 12500/12500. The positive reviews are regarded as the positive class in our experiment. 

\textbf{Mnist} is a simple image dataset. It consists of $28\times28$ grayscale images. There are 10 classes corresponding to digits from 0 to 9. The number of train/test samples per class is almost 6000/1000. We let the images with label 2 as the positive class and the rest images as the negative class in our experiment.

\textbf{Fashion-Mnist} is a new dataset comprising of $28\times28$ grayscale images of 70000 fashion products with 10 categories. It is designed to serve as a direct drop-in replacement for the original Mnist dataset. The training dataset has 6000 images per class while the test dataset has 1000 images per class. To evaluate our algorithm on various scales of datasets, two simulated data sets of different sizes are extracted from this dataset. The first one chooses the images labeled by 0,2 (T-Shirt, Pullover) as the positive class and the images labeled by 1,3 (Trouser, Dress) as the negative class. The second one chooses the images labeled by 4,5,6 (Coat, Sandal, Shirt) as the positive class and the images labeled by 7,8,9 (Sneaker, Bag, Ankle boot) as the negative class.

\textbf{Cifar-10} is a more complex image dataset than Fashion-Mnist. It contains 32x32 color images with 10 classes of natural objects. The standard train/test split for each class is 5000/1000. There are two simulated data sets of different sizes are extracted from this dataset. The first one chooses the images labeled by 1 (automobile) as the positive class and the images labeled by 3,4,5,6 (cat, deer, dog, frog) as the negative class.The other one takes  the images labeled by 7 (horse) as the positive class and the images labeled by 8,9 (ship, truck) as the negative class. 

The training dataset with different imbalance levels are obtained by reducing the number of positive class to $\rho\times N$ where $N$ is the total number of negative class and $\rho$  is imbalanced ratio of dataset. The detail description of experiment dataset is shown in Table \ref{tbl1}.

\subsection{Network Architecture}
We use deep neural network to learn the feature representation from the imbalanced and high dimensional datasets. For the compared algorithms, the network architecture used for text (IMDB) dataset has a embedding layer and two fully connected layers and a softmax output layer. The detailed parameters are given in Table \ref{tbl2}. The network architecture that is used for image (Mnist, Fashion-Mnist,Cifar-10) classification has two convolution layers and two fully connected layers and a softmax output layer. Its detailed parameters are given in Table \ref{tbl3}. For our model, the $Q$ network architecture is similar to the network structure of compared algorithms, but the final softmax output layer is removed because it does not need to scale the $Q$ value of different actions between 0 and 1.

\begin{table}[]
\caption{Network architecture used for text dataset}
\label{tbl2}
\centering
\begin{tabular}{lcc}
\toprule
Layer          & Input    & Output   \\
\midrule
Embedding      & 500      & (500,64) \\
Flatten        & (500,64) & (32000)  \\
FullyConnected & (32000)  & 250      \\
ReLU           & -        & -        \\
FullyConnected & 250      & 2        \\
Softmax        & 2        & 2        \\
\bottomrule
\end{tabular}
\end{table}

\begin{table}[]
\caption{Network architecture used for image dataset}
\label{tbl3}
\centering
\begin{tabular}{lccccc}
\toprule
Layer          & Width  & Height & Depth      & Kernel size & Stride \\
\midrule
Input          & 28(32) & 28(32) & 1(3)       & -           & -      \\
Convolution    & 28(32) & 28(32) & 32         & 5           & 1      \\
ReLU           & 28(32) & 28(32) & 32         & -           & -      \\
MaxPooling     & 14(16) & 14(16) & 32         & 2           & 2      \\
Convolution    & 14(16) & 14(16) & 32         & 5           & 1      \\
ReLU           & 14(16) & 14(16) & 32         & -           & -      \\
MaxPooling     & 7(8)   & 7(8)   & 32         & 2           & 2      \\
Flatten        & 1      & 1      & 1568(2048) & -           & -      \\
FullyConnected & 1      & 1      & 256        & -           & -      \\
ReLU           & 1      & 1      & 256        & -           & -      \\
FullyConnected & 1      & 1      & 2          & -           & -      \\
Softmax        & 1      & 1      & 2          & -           & -       \\
\bottomrule
\end{tabular}
\end{table}

\begin{table}[]
\caption{Experiment results on balanced datasets}
\label{tbl5}
\centering
\begin{tabular}{ccccc}
\toprule
\multirow{2}{*}{\begin{tabular}[c]{@{}c@{}}Dataset\\ (balanced)\end{tabular}} & \multicolumn{2}{c}{G-mean}   & \multicolumn{2}{c}{F-measure} \\
\cline{2-5}
                                                                             & DNN & DQNimb & DNN  & DQNimb  \\
\midrule
IMDB                                                                         & \textbf{0.864}         & \textbf{0.864}        & 0.863         & \textbf{0.865}         \\
Cifar-10(1)                                                                  & 0.962         & \textbf{0.967}        & 0.941         & \textbf{0.950}         \\
Cifar-10(2)                                                                  & 0.959         & \textbf{0.963}        & 0.946         & \textbf{0.952}         \\
Fashion-Mnist(1)                                                                & 0.978         & \textbf{0.984}        & 0.978         & \textbf{0.984}         \\
Fashion-Mnist(2)                                                                & 0.990         & \textbf{0.991}        & 0.990         & \textbf{0.991}         \\
Mnist                                                                        & 0.995         & \textbf{0.997}        & 0.985         & \textbf{0.992}      \\
\bottomrule
\end{tabular}
\end{table}
\begin{table*}[htb]
\caption{G-mean score of experiment results}
\centering
\label{tbl4}
\begin{tabular}{ccccccccc}
\toprule
Dataset                         & \tabincell{c}{Imbalance\\ratio $\rho$} & \tabincell{c}{DQNimb\\(Ours)}                                & \tabincell{c}{Baseline\\(DNN)} & \tabincell{c}{MFE loss\\(MFE)}                                   & \tabincell{c}{Over-sampling\\(ROS)}                         & \tabincell{c}{Under-sampling\\(RUS)}                        & \tabincell{c}{Cost-sensitive\\(CSM)}                        & \tabincell{c}{Threshold-Adjustment\\(DTA)}                  \\
\toprule
                                & 10\%                   & {\color[HTML]{FE0000} \textbf{0.820}} & 0.548    & 0.687                                 & 0.681                                 & 0.740                                 & {\color[HTML]{3166FF} \textbf{0.743}} & 0.678                                 \\
                                & 5\%                    & {\color[HTML]{FE0000} \textbf{0.781}} & 0.299    & 0.589                                 & 0.632                                 & 0.622                                 & {\color[HTML]{3166FF} \textbf{0.696}} & 0.599                                 \\
\multirow{-3}{*}{IMDB}          & 2\%                    & {\color[HTML]{FE0000} \textbf{0.682}} & 0.034    & 0.351                                 & 0.343                                 & 0.510                                 & {\color[HTML]{3166FF} \textbf{0.559}} & 0.355                                 \\
\midrule
                                & 4\%                    & {\color[HTML]{FE0000} \textbf{0.956}} & 0.869    & 0.939                                 & {\color[HTML]{3166FF} \textbf{0.947}} & 0.945                                 & 0.944                                 & 0.946                                 \\
                                & 2\%                    & {\color[HTML]{FE0000} \textbf{0.941}} & 0.824    & 0.908                                 & 0.925                                 & {\color[HTML]{3166FF} \textbf{0.929}} & 0.922                                 & 0.928                                 \\
                                & 1\%                    & {\color[HTML]{FE0000} \textbf{0.917}} & 0.730    & 0.859                                 & 0.897                                 & 0.896                                 & 0.884                                 & {\color[HTML]{3166FF} \textbf{0.912}} \\
\multirow{-4}{*}{Cifar-10(1)}   & 0.5\%                  & {\color[HTML]{3166FF} \textbf{0.890}} & 0.579    & 0.759                                 & 0.838                                 & 0.866                                 & 0.853                                 & {\color[HTML]{FE0000} \textbf{0.901}} \\
\midrule
                                & 4\%                    & {\color[HTML]{FE0000} \textbf{0.925}} & 0.815    & 0.882                                 & 0.904                                 & 0.906                                 & 0.911                                 & {\color[HTML]{3166FF} \textbf{0.915}} \\
                                & 2\%                    & {\color[HTML]{FE0000} \textbf{0.917}} & 0.758    & 0.852                                 & 0.894                                 & 0.887                                 & 0.886                                 & {\color[HTML]{3166FF} \textbf{0.908}} \\
                                & 1\%                    & {\color[HTML]{FE0000} \textbf{0.883}} & 0.677    & 0.769                                 & 0.854                                 & 0.859                                 & 0.850                                 & {\color[HTML]{3166FF} \textbf{0.873}} \\
\multirow{-4}{*}{Cifar-10(2)}   & 0.5\%                  & {\color[HTML]{FE0000} \textbf{0.829}} & 0.513    & 0.693                                 & 0.792                                 & {\color[HTML]{3166FF} \textbf{0.822}} & 0.816                                 & 0.821                                 \\
\midrule
                                & 4\%                    & {\color[HTML]{FE0000} \textbf{0.971}} & 0.921    & 0.960                                 & 0.962                                 & 0.957                                 & {\color[HTML]{3166FF} \textbf{0.964}} & {\color[HTML]{3166FF} \textbf{0.964}} \\
                                & 2\%                    & {\color[HTML]{FE0000} \textbf{0.966}} & 0.885    & 0.947                                 & 0.957                                 & 0.953                                 & 0.956                                 & {\color[HTML]{3166FF} \textbf{0.962}} \\
                                & 1\%                    & {\color[HTML]{FE0000} \textbf{0.959}} & 0.853    & 0.934                                 & 0.948                                 & 0.943                                 & 0.946                                 & {\color[HTML]{3166FF} \textbf{0.952}} \\
\multirow{-4}{*}{Fashion-Mnist(1)} & 0.5\%                  & {\color[HTML]{FE0000} \textbf{0.950}} & 0.757    & 0.901                                 & 0.927                                 & 0.934                                 & 0.924                                 & {\color[HTML]{3166FF} \textbf{0.944}} \\
\midrule
                                & 4\%                    & {\color[HTML]{FE0000} \textbf{0.985}} & 0.951    & 0.968                                 & 0.972                                 & 0.967                                 & 0.973 
                                & {\color[HTML]{3166FF}\textbf{0.977}} \\
                                & 2\%                    & {\color[HTML]{FE0000} \textbf{0.982}} & 0.926    & 0.960                                 & 0.963                                 & 0.956                                 & 0.966                                 & {\color[HTML]{3166FF} \textbf{0.970}} \\
                                & 1\%                    & {\color[HTML]{FE0000} \textbf{0.979}} & 0.872    & 0.940                                 & 0.949                                 & 0.946                                 & 0.958                                 & {\color[HTML]{3166FF} \textbf{0.962}} \\
\multirow{-4}{*}{Fashion-Mnist(2)} & 0.5\%                  & {\color[HTML]{FE0000} \textbf{0.972}} & 0.821    & 0.912                                 & 0.935                                 & 0.937                                 & 0.950                                 & {\color[HTML]{3166FF} \textbf{0.953}} \\
\midrule
                                & 1\%                    & {\color[HTML]{FE0000} \textbf{0.991}} & 0.967    & {\color[HTML]{3166FF} \textbf{0.982}} & 0.981                                 & 0.978                                 & {\color[HTML]{3166FF} \textbf{0.982}} & 0.978                                 \\
                                & 0.2\%                  & {\color[HTML]{FE0000} \textbf{0.983}} & 0.923    & 0.949                                 & 0.944                                 & 0.953                                 & 0.951                                 & {\color[HTML]{3166FF} \textbf{0.961}} \\
                                & 0.1\%                  & {\color[HTML]{FE0000} \textbf{0.968}} & 0.856    & 0.921                                 & 0.911                                 & 0.929                                 & {\color[HTML]{3166FF} \textbf{0.942}}                                 & 0.937 \\
\multirow{-4}{*}{Mnist}         & 0.05\%                 & {\color[HTML]{FE0000} \textbf{0.941}} & 0.694    & 0.842                                 & 0.858                                 & 0.907 & {\color[HTML]{3166FF} \textbf{0.921}}                                 & 0.916         \\
\bottomrule
\multicolumn{9}{l}{The $1^{st}/2^{nd}$ best results are indicated in red/blue.}
\end{tabular}
\end{table*}

\begin{figure*}[htb]
    \centering
    \includegraphics[width=19cm, height=9cm]{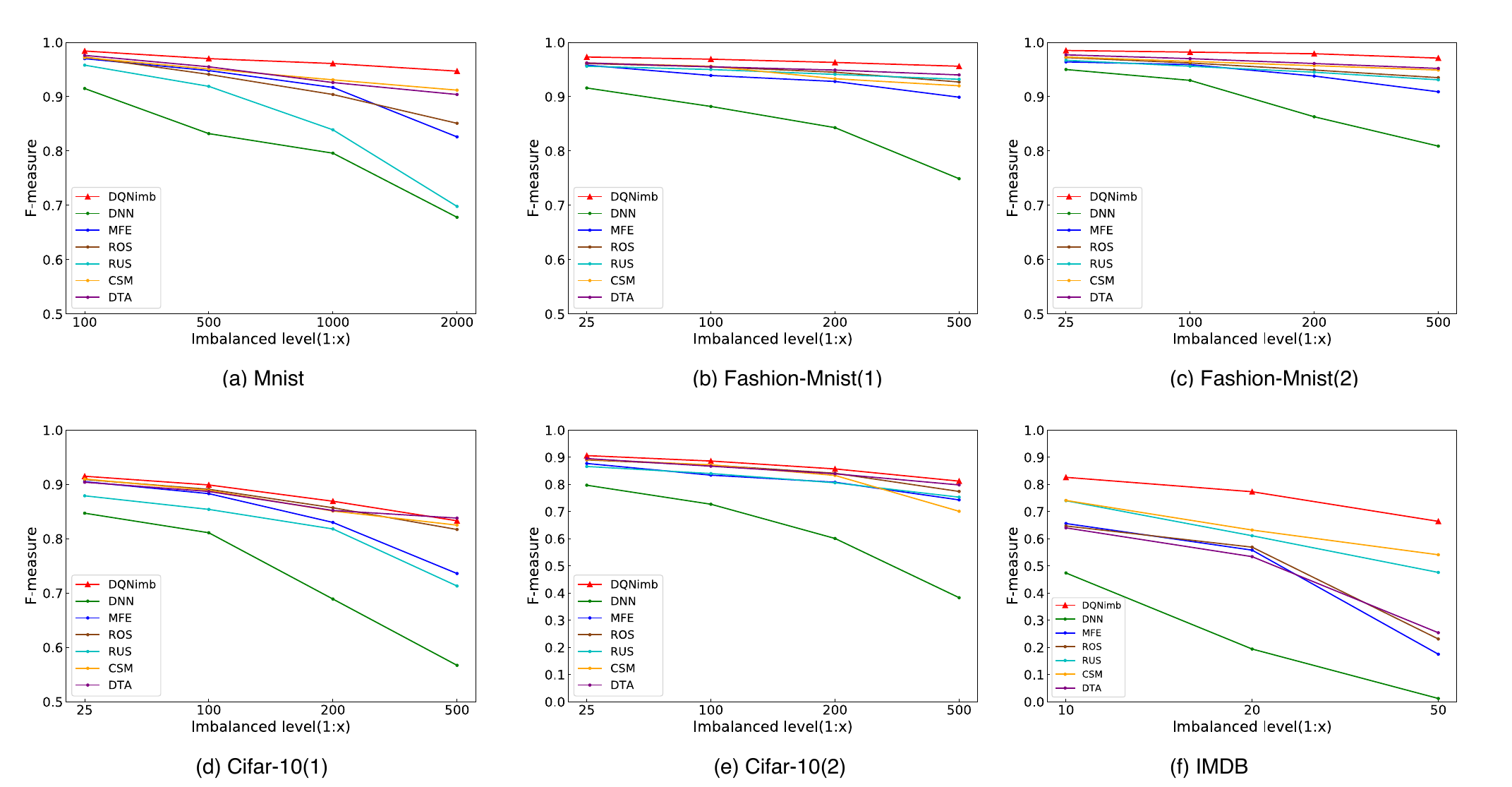}
    \caption{Comparison of methods with respect to F-measure score on different datasets.}
    \label{fig2}
\end{figure*}
\subsection{Parameter Setting}
We use $\epsilon$-greedy policy for DQN based imbalanced classification model in which the probability of exploration $\epsilon$ is linearly attenuated from 1.0 to 0.01. The size of experience replay memory is 50 000 and the interactions between agent and environment are approximately 120 000 steps. The discount factor of immediate reward $\gamma$ is 0.1. Adam algorithm is used to optimize the parameters of $Q$-network and its learning rate is 0.00025. For other algorithms, the optimizer is Adam and its learning rate is 0.0005, the batch size is 64. We randomly select 10\% samples of training data as the verification data and use early stopping technique \cite{bengio2012practical} which monitors the validation loss to train the deep neural network for 100 epochs.

\subsection{Experiment Result}
Before the research of imbalanced data learning, we compare our DQNimb model to the DNN that is a supervised deep learning model in balanced data sets. The experiments were conducted on the six data sets (the imbalance ratio $\rho$ is 1) in Table \ref{tbl1}. The number of positive samples and negative samples are equal, so the reward function of the DQNimb model assigns the same reward or punishment to the positive and negative samples. For fairness and convincing comparisons, the network architecture of the DNN model is the same as the Q network architecture of the DQNimb model. The G-mean scores and F-measure scores of the experimental results are shown in Table \ref{tbl5}. Despite of the different learning mechanisms, that the DQNimb model obtains the optimal classification strategy by maximizing the cumulative rewards in the Markov process, while the DNN gets the optimal network parameters by minimize the cross-entropy loss function, both models demonstrate good performance in experimental results. The G-mean scores and F-measure scores of the DQNimb model are slightly better than those of the DNN model.

Given the number of the negative samples of the imbalanced data set is $N$, we randomly select $\rho\times N$ positive samples according to the imbalance ratio $\rho$, and conducted 6 experiments. We report the G-mean scores of our method and the other methods on the different imbalanced data sets in Table \ref{tbl4}. Each training was repeated 5 times on the same data set. The results of data sampling methods, cost-sensitive learning methods and threshold adjustment method are much better than DNN model in imbalanced classification problems, however, our model DQNimb achieves an outstanding performance with an overwhelming superiority. In the IMDB text dataset, G-mean score of our method DQNimb are normally 7.7\% higher than the second-ranked method CSM, and are even 12.3\% higher when the imbalance ratio is 2\%.

We report the F-measure scores of different algorithms in Fig.\ref{fig2}. With the increase of data imbalance level, the F-measure scores of each algorithm show a significant decline. The DNN model suffers the most serious declination, that is, DNN can hardly identify any minority class sample when the data distribution is extremely imbalanced. Meanwhile, our model DQNimb enjoys the smallest decrease because our algorithm possesses both the advantages of the data level models and the algorithmic level models. In the data level, our model DQNimb has an experience replay memory of storing interactive data during the learning process. When the model misclassifies a positive sample, the current episode will be terminated, this can alleviate the skewed distribution of the samples in the experience replay memory. In the algorithmic level, the DQNimb model gives a higher reward or penalty for positive samples, which raises the attention to the samples in minority class and increases the probabilities that positive samples are correctly identified.

\subsection{Exploration On Reward Function}\label{sec:IRF}
\begin{figure}[htbp]
    \centering
    \includegraphics[width=9cm, height=12cm]{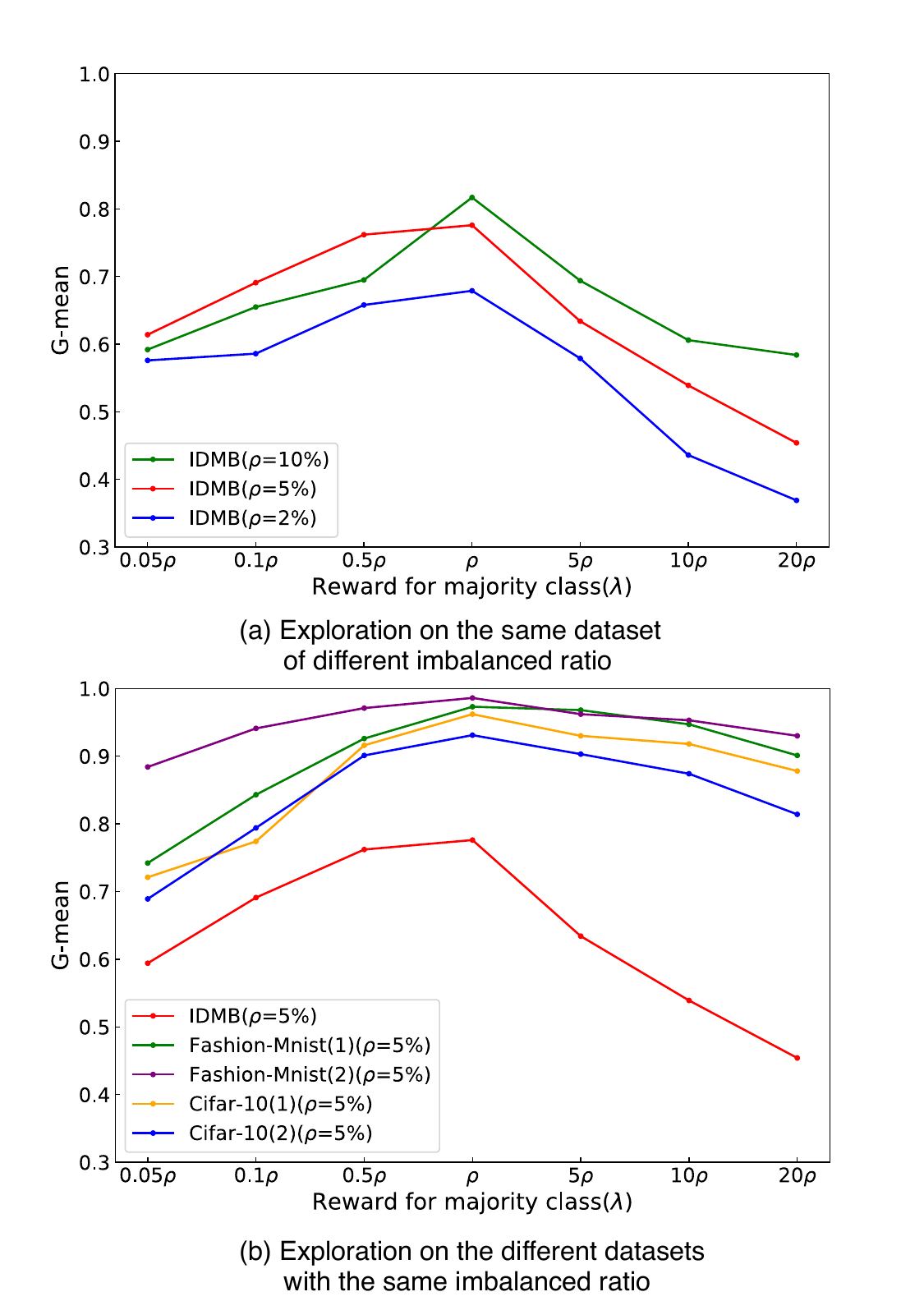}
    \caption{Different rewards for majority class to find the optimal reward function.}
    \label{fig3}
\end{figure}

Reward function is used to evaluate the value of actions performed by agent and inspires it to work toward to the goal. In DQNimb model, the reward of minority class is 1 and the reward of majority class is $\lambda$. In above experiments, we let $\lambda=\rho$. To study the effect of different values of $\lambda$ on the classification performance, we test values of $\lambda\in\lbrace 0.05\rho,0.1\rho,0.5\rho,\rho,5\rho,10\rho,20\rho\rbrace$. The experimental results are shown in Fig.\ref{fig3}. 

In the same dataset of distinct imbalanced degree, the model performs best when the reward of majority class $\lambda$ is equal to the imbalanced ratio $\rho$. In different datasets with the same imbalanced ratio, the closer the reward of majority class $\lambda$ is to $\rho$, the better the classification performance of model is, that is, the different values of
$\lambda$ can adjust the impact of majority samples on classification performance. Increasing or decreasing the value of $\lambda=\rho$ will break the balance of the second item and the third item in \eqref{eq.16} and lead to a poor classification performance.

\section{Conclusion}
This paper introduces a novel model for imbalanced classification using a deep reinforcement learning. The model formulates the classification problem as a sequential decision-making process (ICMDP), in which the environment returns a high reward for minority class sample but a low reward for majority class sample, and the episode will be terminated when the agent misclassifies the minority class sample. We use deep Q learning algorithm to find the optimal classification policy for ICMDP, and theoretically analyze the impact of the specific reward function on the loss function of Q network when training. The effect of the two types of samples on the loss function can be balanced by reducing the reward value the agent receives from the majority samples. Experiments showed that our model's classification performance in imbalanced data sets is better than other imbalanced classification methods, especially in text data sets and extremely imbalanced data sets. In the future work, we will apply improved deep reinforcement learning algorithms to our model, and explore the design of reward function and the establishment of learning environment for classification in imbalanced multi-class data sets.

\bibliographystyle{IEEEtran}
\bibliography{drl_imb_classification}

\end{document}